\documentclass[letterpaper, 10 pt, conference]{ieeeconf}  

\IEEEoverridecommandlockouts                              

\usepackage{amssymb}  

\usepackage{siunitx}
\usepackage{tabulary}
\usepackage{graphicx}
\usepackage{subcaption}
\usepackage{threeparttable}
\usepackage{flushend}

\makeatletter
\let\NAT@parse\undefined
\makeatother
\usepackage{hyperref}  

\title{\LARGE \bf
Large-scale Ridesharing DARP Instances 
Based on Real Travel Demand
}

\author{David Fiedler$^{1}$ and Jan Mrkos$^{2}$
\thanks{*This work has been supported by the EU OP RDE funded project Research Center for Informatics; reg. No.: \texttt{CZ.02.1.01/0.0/0.0/16\_019/0000765} and by the Grant Agency of the Czech Technical University in Prague, grant No. \texttt{SGS22/168/OHK3/3T/13.}}
\thanks{$^{1}$David Fiedler is with the Faculty of Electrical Engineering,
        Czech Technical University in Prague, Karlovo náměstí 13, Prague 2, 121 35, Czechia
        {\tt\small david.fiedler@fel.cvut.cz}}%
\thanks{$^{2}$Jan Mrkos is with the Faculty of Electrical Engineering,
        Czech Technical University in Prague, Karlovo náměstí 13, Prague 2, 121 35, Czechia
        {\tt\small jan.mrkos@fel.cvut.cz}}%
}

\begin{document}

\maketitle
\thispagestyle{empty}
\pagestyle{empty}

\begin{abstract}
\setcounter{footnote}{2}
Accurately predicting the real-life performance of algorithms solving the Dial-a-Ride Problem (DARP) in the context of Mobility on Demand (MoD) systems with ridesharing requires evaluating them on representative instances. 
However, the benchmarking of state-of-the-art DARP solution methods has been limited to small, artificial instances or outdated non-public instances, hindering direct comparisons. 
With the rise of large MoD systems and the availability of open travel demand datasets for many US cities, there is now an opportunity to evaluate these algorithms on standardized, realistic, and representative instances. 
Despite the significant challenges involved in processing obfuscated and diverse datasets, we have developed a methodology using which we have created a  comprehensive set of large-scale demand instances based on real-world data\footnote{All instances, results, and source code are available through the instance repository: \url{https://github.com/aicenter/Ridesharing_DARP_instances}}. 
These instances cover diverse use cases, one of which is demonstrated in an evaluation of two established DARP methods: the insertion heuristic and optimal vehicle-group assignment method. 
We publish the full results of both methods in a standardized format.
The results show significant differences between areas in all measured quantities, emphasizing the importance of evaluating methods across different cities.
\end{abstract}

\section{INTRODUCTION}
Dial-a-ride-problem (DARP) is a traditional problem from operational research~\cite{tothVehicleRoutingProblems2014}.
For more than five decades, researchers come up with new methods for solving DARP, which they evaluate on various problem instances.
Most of the articles from the last two decades use randomly generated DARP instances and Euclidean disctances~\cite{cordeauTabuSearchHeuristic2003, cordeauBranchandCutAlgorithmDialaRide2006, ropkeAdaptiveLargeNeighborhood2006, ropkeBranchCutPrice2009, parraghIntroducingHeterogeneousUsers2011, kirchlerGranularTabuSearch2013, gschwindAdaptiveLargeNeighborhood2019, masmoudiHybridAdaptiveLarge2020, hamDialarideProblemMixed2021}.
These instances usually contain just tens of requests and vehicles, in line with the original application for transportation of people with disabilities.

Recently, a new vehicle scheduling problem arose with the emergence of large-scale mobility-on-demand (MoD), mostly operated by transportation network companies.
Some of the MoD systems use ridesharing, i.e., serving multiple travel requests at once by one vehicle.
With ridesharing, the dispatching problem of the MoD system is essentially a DARP.
However, the problem characteristics are very different compared to classical DARP.
First, for an MoD system, the instances usually cover shorter time periods, especially the ones covering online ridesharing.
Thus, the limits on the maximum route duration present in the classical DARP formulation are not necessary.
Second, the MoD instances are much larger than the traditional DARP instances.
In New York City, for example, there can be as many as \num{100000} active taxis during peak traffic hours, at least \emph{three orders of magnitude} more than in the largest traditional DARP instances~\cite{nyctaxilimousinecommission2018Factbook2018}.
Finally, the time windows and maximum delays are much shorter as there are higher requirements regarding the quality of service.

At the same time, the new trend of public open data has allowed researchers to create problem instances from real travel demand data.
Because of this, the DARP solution methods for MoD systems are usually evaluated on large-scale instances based on real travel demand datasets.
Creating large-scale MoD problem instances using real demand data offers several advantages. Firstly, it provides a realistic representation of both the scale and distribution of the demand. This ensures that the instances closely resemble real-world scenarios. Secondly, incorporating the temporal changes in demand into the problem instances allows for a more accurate reflection of dynamic patterns and variations in service requirements over time. Lastly, by utilizing road network travel times instead of relying solely on Euclidean distance, the computed travel cost aligns more closely with actual transportation conditions, enhancing the realism of the instances.

However, large realistic problem instances present new challenges.
Instead of simple random instance generation, we must process and transform the travel demand datasets into the instance data.
Moreover, in recent years, data providers started to obfuscate the published travel demand datasets to protect the privacy of users and drivers.
This means that the request origin and destination locations and, sometimes, even request times are not specified precisely. 
Instead, locations are binned into zones and times into time intervals.
Thus, we must first generate trip locations and times to create instances from obfuscated data. 
The absence of a standardized format for open data across various cities further complicates the situation; a procedure for creating instances from one dataset cannot be applied to other datasets without significant changes.
Apart from demand data, we require other datasets for every location where we wish to create a problem instance. At the very least, researchers must process the road network data and, possibly, some dataset of travel speeds to obtain realistic travel times. 

To avoid these pitfalls, many works use the Manhattan travel demand datasets published before the introduction of privacy protection rules in 2014. 
However, the Manhattan dataset has a highly distinctive geography and unusually high demand density. Moreover, due to the 2014 cutoff, new phenomena, such as the displacement of traditional taxi services by transportation network companies or the COVID pandemic, are not captured in datasets based on the pre-2014 Manhattan demand data. 
This is important since many new methods are evaluated \emph{only} using the Manhattan demand dataset~\cite{santiQuantifyingBenefitsVehicle2014, alonso-moraOndemandHighcapacityRidesharing2017, vazifehAddressingMinimumFleet2018, wallarOptimizingVehicleDistributions2019b, beirigoBusinessClassAutonomous2022, fiedlerLargescaleOnlineRidesharing2022}. 
As a result, the field of DARP research runs the risk of overfitting to specific problem instances that lack generalizability to different locations or current travel patterns. This can result in misleading conclusions and unrealistic expectations when evaluating DARP solution methods. 

At first glance, it may appear that using old Manhattan instances at least provides the standardization we know from the classical DARP instances~\cite{cordeauTabuSearchHeuristic2003,cordeauBranchandCutAlgorithmDialaRide2006,ropkeModelsBranchandcutAlgorithms2007}.
The ability to compare methods to previously established ones without the need for re-implementation is a crucial factor for productivity.
However, there are no publicly available Manhattan instances. 
Instead, every work presents its own instances based on the Manhattan dataset without publishing them; thus, their results are hardly comparable or replicable.

In this work, we present new large-scale DARP instances based on real travel demand data from three cities: New York, Chicago, and Washington, DC.
The instances are based on recently collected data, capturing the latest developments in the MoD field.
Moreover, we present a detailed description of the process of creating these instances, supported by the source code needed to create them.
Finally, we evaluated the instances using two known methods for solving DARP to demonstrate the advantages of using multiple areas and instance configurations.
The methods used are 1) a simple construction heuristic: the insertion heuristic, 
and 2) an optimal solution method: the vehicle-group assignment method~\cite{alonso-moraOndemandHighcapacityRidesharing2017}.
Together with the proposed instances, we distribute full solutions computed by these methods in the instance repository~(see footnote 3 on page 1).

\section{METHODOLOGY}
The instances we present are large ridesharing DARP instances with several differences compared to classical DARP instances.
Each instance contains a) demand: a list of requests for transportation between 2 points, b) a set of vehicles represented by their starting locations, and c) a model of travel time between any two locations.

We denote a set of all locations (vertices) in the road network as $ L $.
Each travel request is a 3-tuple \(r = (o, d, t)\) consisting of the origin location $ o \in L $, destination location $ d \in L $, and desired pickup time, respectively.
Each vehicle is a 2-tuple \(v = (s, c)\), where \(s \in L \) is the vehicle starting location and \(c\) its transport capacity (available seats for passengers). 
Formally, each proposed DARP instance is then a 4-tuple \((R, V, f_{t}, \Delta)\) with:
\begin{itemize}
\item \(R = [r_1, r_2, \ldots, r_n]\), a list of \(n\) travel requests ordered by their desired pickup time,
\item \(V = \{v_1, v_2, \ldots, r_m\}\) a set of available vehicles,
\item \(f_t(l, l')\) is a function that models the travel time times (in seconds) between any $ l, l' \in L $.
\item \(\Delta\) is the maximum delay parameter of the instance. It determines the maximum extra time to the requested trip compared to direct and immediate travel between the origin and destination location.
\end{itemize}

\begin{figure*}
    \centering
    \includegraphics[width=\linewidth]{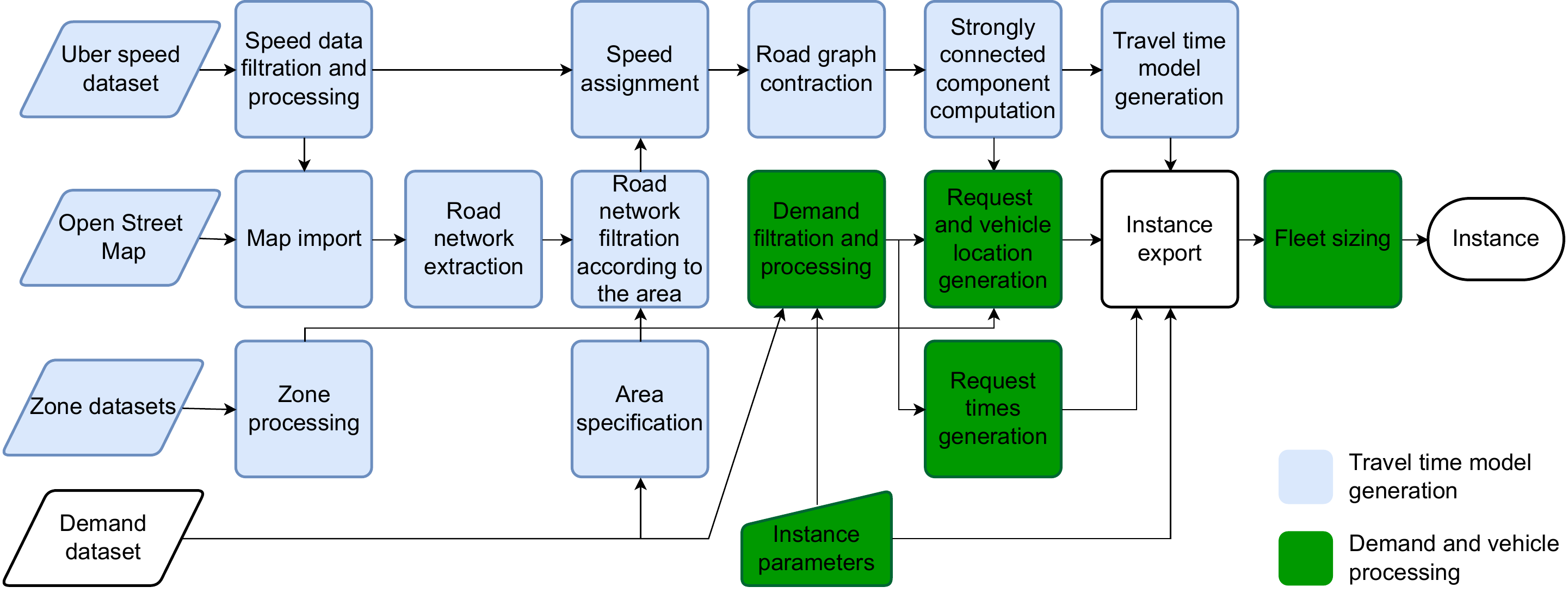}
    \caption{Flow chart of the instance creation process. 
    }
    \label{fig:flow_chart}
\end{figure*}

The rest of this section describes the instance creation methodology from a high-level perspective. 
For a detailed description of the methods used and their implementation, you can visit the instance repository.
We present the travel time computation for the travel time model \(f_t\) in Section~\ref{sec:travel_times}.
We use open travel data specific to each city to generate the demand \(R\) and vehicles \(V\). 
This is described in Section~\ref{sec:demand_and_vehicles}.
Finally, in Section~\ref{sec:differences}, we explain the differences between classical DARP instances and the proposed instances for ridesharing DARP.
The flow chart covering the whole process of creating an instance is in Figure~\ref{fig:flow_chart}. 

\subsection{Computing Travel Time}
\label{sec:travel_times}
In classical DARP instances, the request origin and destination locations, as well as the start locations for vehicles, are coordinates in Euclidean space.
The travel time between any two points is then the distance between these points.
However, real travel times are much more complex; 
real travel times are not symmetric and cannot be calculated purely from origin-destination coordinates.
To approximate the travel time, we use a road network with assigned speeds for each road segment and compute the shortest path between locations in the road network.

We base the travel time model on two datasets: the OpenStreetMap\footnote{\url{https://www.openstreetmap.org/}} for the road network, and the UberMovement dataset\footnote{\url{https://movement.uber.com/}} for speeds.
The shortest-path computations are expensive, which is exacerbated by the fact that most methods for solving DARP need to compute travel times frequently.
Therefore, the travel times are usually precomputed in a distance matrix to reduce the overhead.
However, to keep the size of the matrix manageable, we need to discretize the road network to limit the number of origin-destination pairs. In this work, we use crossroads as locations.

The travel time model for each area was created through the steps shown in light blue in Figure~\ref{fig:flow_chart}. 
The crucial parts of the process are:
\begin{enumerate}
    \item \textbf{Speed data preparation} for the area, date, and time of interest.

    \item \textbf{Area specification}: the instance area is specified as a convex shape such that all the demand in the instances we plan to generate lies inside this shape,
    
    \item \textbf{Import of the selected map} from the OpenStreetMap dataset according to the map state at the time when the speed dataset was created,  

    \item \textbf{Filtration of the map} so that it contains only road network,

    \item \textbf{Speed assignment} to roads according to the speed dataset,

    \item \textbf{Road graph contraction} by elimination of all nodes that are not intersections.

    \item \textbf{Largest strongly-connected component} selection to remove unreachable "islands" either real or artifacts of map filtration.

\end{enumerate}
After the largest strongly-connected component computation, the road network processing is finished, and we can produce the travel time model \(f_t\). 

\subsection{Demand and Vehicles Processing}
\label{sec:demand_and_vehicles}
This section provides a brief description of the process for transforming a travel demand dataset into demand \(R\) and vehicle data \(V\) for the DARP instance. 

The initial step in the creation of the demand component \(R\) involves selecting the date, start time, and end time. These parameters are then used to extract the relevant records from the travel demand dataset for a specific city.

As previously mentioned, the travel demand data are obfuscated by all data providers after 2014 due to privacy concerns.
Consequently, to generate the instance, we must devise a method for generating the demand locations and, in some cases, the demand origin time. 
In both cases, we sample a uniform distribution across nodes in the processed road network (Section \ref{sec:travel_times}) within the designated zone specified in the dataset and across the dataset's pickup time interval, respectively.

Finally, we also need to generate vehicles and determine their starting positions.
To accomplish this, we sample from the demand dataset at a specified time before the instance start time. 
This approach ensures that we obtain realistic vehicle positions, as these vehicles previously serviced the trips in the travel demand datasets. 
The vehicle fleet size in each instance is chosen as the lowest number of randomly selected vehicles that can service all requests when solving by insertion heuristic. This number is then increased by \(5\%\) to introduce a buffer for cases where the insertion heuristic would find the unique optimal solution.

\subsection{Difference between Ridesharing and Classical Instances}
\label{sec:differences}
Although ridesharing DARP instances are formally almost the same as the classical ones, there are some differences.
The following distinctions are based on the existing ridesharing literature\cite{alonso-moraOndemandHighcapacityRidesharing2017,fiedlerLargescaleOnlineRidesharing2022}.
First, the vehicles do not start from a single depot.
Vehicles must be distributed throughout the area to serve the demand with the shortest possible delay while supporting large areas.
Therefore, we use a general formulation where each vehicle has its own initial position.
Second, the ridesharing instance configuration has no maximum route time and no maximum passenger ride time.
The max route time in the classical DARP instances represents the maximum time a driver can work without pause imposed by contractor law. 
These constraints are unnecessary for the ridesharing instances since they cover much shorter time horizons than the classical DARP instances.
Instead of the maximum passenger's ride time, we use a maximum delay: a constant time limit to an extra service time compared to the direct origin-destination travel time.

Besides the formal differences, there are even more impactful changes in the instance configurations (The specific configurations we used are described in the following section, see Table \ref{tab:instances} for examples of classical and proposed instances).
First, the time windows are much shorter.
The maximum allowed delays in ridesharing instances are in minutes, while in classical instances, it is usually in hours.
This change, caused by the limited time flexibility of MoD users, significantly reduces the combinatorial complexity of the problem, as was demonstrated in the literature~\cite{fiedlerLargescaleOnlineRidesharing2022}.
On the other hand, the instance size of the proposed realistic ridesharing instances is many orders of magnitude larger since they are created from historical travel demand.
This translates into much higher computational complexity. Thus, the computational complexity of the proposed instances has a different source than in the classical DARP instances.

\section{DARP INSTANCES}
\begin{table}
\caption{Comparison of selected proposed instances and classical DARP instances}
\label{tab:instances}
\begin{threeparttable}
    \centering
    \setlength\extrarowheight{2pt}
    \setlength{\tabcolsep}{2pt}
    \begin{tabulary}{\linewidth}{llrrCL}
        Instances & Duration & Requests & Vehicles & Mean Trip Dur.  $\pm$ SD [min] & Time Window [min] \\
        \hline
        DC & \SI{30}{s} & \num{4} & \num{18} & 25.0$\pm$7.5 & 3, 5, 10$^c$ \\
        DC & \SI{15}{min} & \num{163} & \num{121} & 15.8$\pm$8.7 & 3, 5, 10$^c$ \\
        DC & \SI{16}{h} & \num{3 297} & \num{218} & 16.1$\pm$8.3 & 3, 5, 10$^c$ \\
        Chicago & \SI{30}{s} & \num{5} & \num{7} & 7.3$\pm$4.0 & 3, 5, 10$^c$ \\
        Chicago & \SI{15}{min} & \num{274} & \num{198} & 13.4$\pm$15.1 & 3, 5, 10$^c$ \\
        Chicago & \SI{16}{h} & \num{3 794} & \num{388} & 17.8$\pm$17.9 & 3, 5, 10$^c$ \\
        Man. & \SI{30}{s} & \num{140} & \num{124} & 7.9$\pm$3.7 & 3, 5, 10$^c$ \\
        Man. & \SI{15}{min} & \num{5 113} & \num{1 672} & 7.7$\pm$3.8 & 3, 5, 10$^c$ \\
        Man. & \SI{16}{h} & \num{90 533} & \num{2 011} & 8.0$\pm$4.0 & 3, 5, 10$^c$ \\
        NYC & \SI{3}{s} & \num{329} & \num{336} & 10.3$\pm$6.2 & 3, 5, 10$^c$ \\
        NYC & \SI{15}{min} & \num{10567} & \num{5 085} & 9.9$\pm$6.6 & 3, 5, 10$^c$ \\
        NYC & \SI{16}{h} & \num{198727} & \num{6 461} & 10.5$\pm$6.7 & 3, 5, 10$^c$ \\
        \hline
        classic\cite{cordeauBranchandCutAlgorithmDialaRide2006}$^a$ & \hspace{0.1em} \SI{24}{h} & 16-96 & 2-8 & 10.6$\pm$4.9 & \SI{15}{min} or \SI{24}{h$^d$} \\
        classic\cite{kirchlerGranularTabuSearch2013}$^b$ & \hspace{0.1em} 
        \SI{24}{h} & 24-144 & 3-13 & 6.8$\pm$3.6 &
        15-\SI{90}{min} or \SI{24}{h$^d$}\\
        \hline
    \end{tabulary}
\begin{tablenotes}
    \item \footnotesize{$^a$48 instances,  $^b$20 instances, $^c$one instance for each, $^d$combined in same instances}
\end{tablenotes}
\end{threeparttable}
\end{table}

\begin{figure}
    \centering
    \begin{subfigure}[b]{1\linewidth}
        \includegraphics[width=\linewidth]{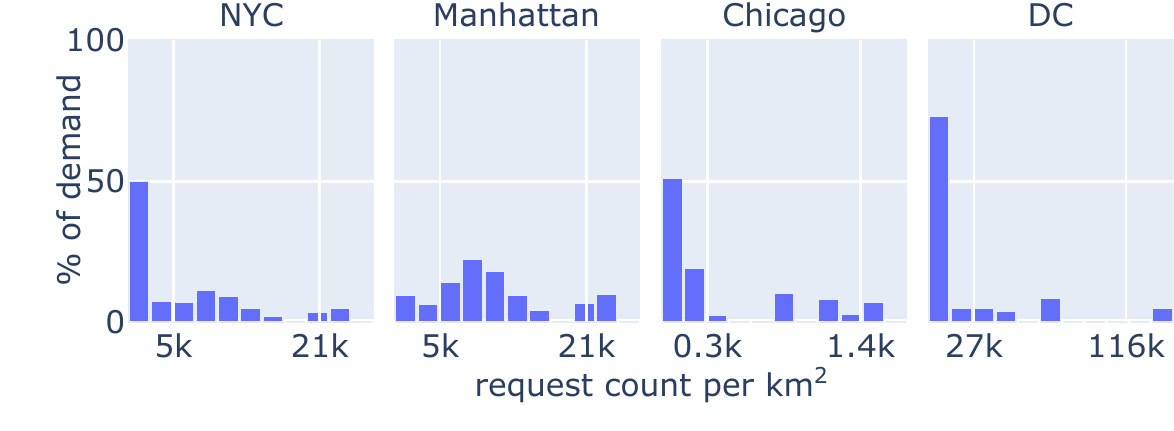}
        \caption{Demand density histogram. The x-axis represents the density of the daily demand in the request's origin zone (Note that it has a different scale for each area).} 
    \end{subfigure}
    \begin{subfigure}[b]{1\linewidth}
        \includegraphics[width=\linewidth]{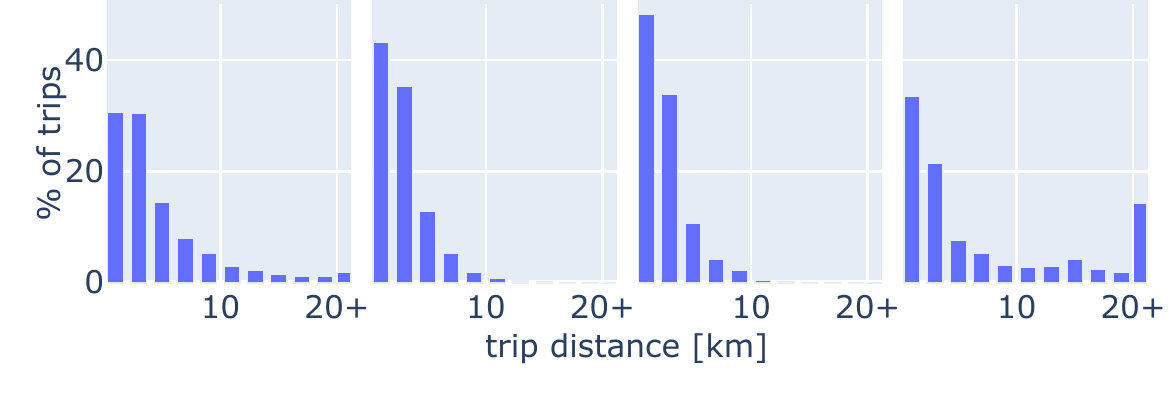}
        \caption{Trip Length Histogram} 
    \end{subfigure}
    \caption{Statistics of the daily demand in each instance area.}
    \label{fig:area_stats}
\end{figure}

We generated instances for four areas: New York City, Manhattan, Chicago, and Washington, DC.
The data sources for those areas are listed in Table~\ref{tab:sources}.
\begin{table}[]
\begin{threeparttable}
    \centering
    \caption{Demand and zone data sources$^a$}
    \label{tab:sources}
    \settowidth\tymin{Request}
    \setlength\extrarowheight{2pt}
    \begin{tabulary}{\linewidth}{LLLLL}
    Area name & Demand data source & Zone data source & Request times & Req/h/km\textsuperscript{2} \\ 
    \hline
    New York City, and Manhattan & NYC taxi and limousine commission & NYC taxi zones & exact & 27, \newline 267 \\
    Chicago & City of Chicago & Census tracts and community areas & generated & 1\\
    Washington DC & City of Washington, DC & Master Address Repository & generated & 4 \\
    \hline
    \end{tabulary}
    \begin{tablenotes}
        \item{$^a$The download links for all data sources can be found in the instance repository}
    \end{tablenotes}
\end{threeparttable}
\end{table}
Each of the generated areas has different characteristics of the travel demand. 
In Figure~\ref{fig:area_stats}, we can see a comparison of the areas using two quantities: demand density and trip length.
Note that the characteristics differ significantly between the areas. 
A typical travel request in Manhattan originates in a zone with about \num{10000} requests per km per day. 
In Chicago, a typical request originates in a zone with a demand density of two orders of magnitude lower.
A similar difference can be seen in the average trip length, which is low in Manhattan and Washington, DC, and high in Chicago.

\begin{figure}
    \centering
    \begin{subfigure}[b]{1\linewidth}
        \includegraphics[width=\linewidth]{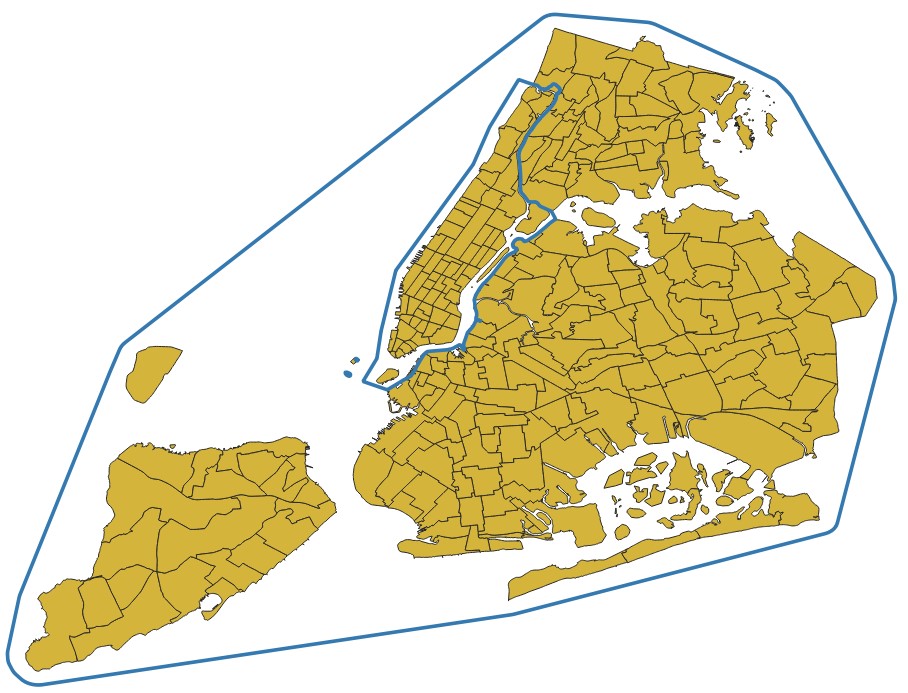}
        \caption{New York City and Manhattan} 
    \end{subfigure}
    \begin{subfigure}[b]{0.49\linewidth}
        \includegraphics[width=\linewidth]{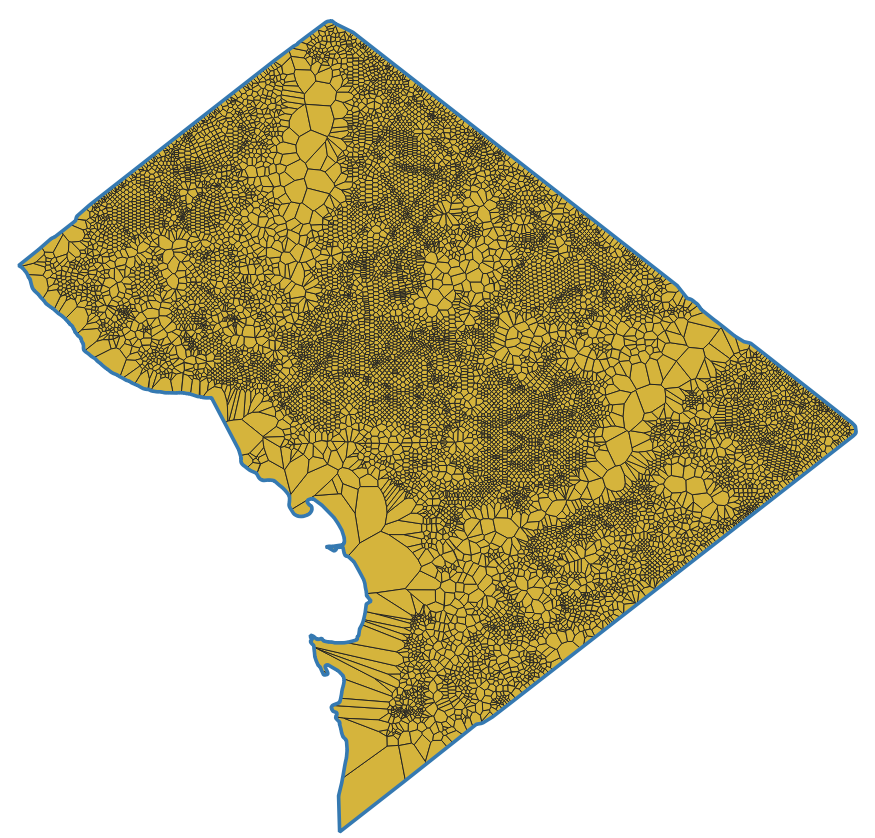}
        \caption{Washington, DC} 
    \end{subfigure}
    \begin{subfigure}[b]{0.49\linewidth}
        \includegraphics[width=\linewidth]{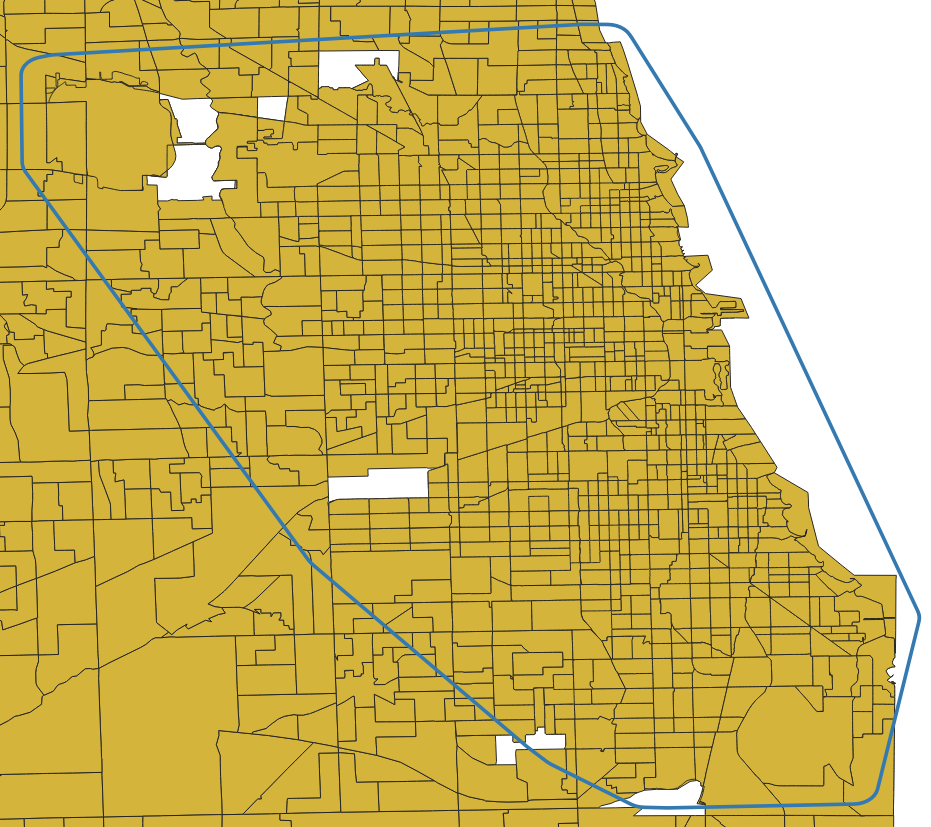}
        \caption{Chicago} 
    \end{subfigure}
    \caption{Area boundaries (blue) and zones (gold)}
    \label{fig:zones}
\end{figure}

For each area, we determined the boundary differently. 
For Manhattan and Washington, DC, we used the administrative boundary.
However, the administrative boundaries of the remaining areas do not match the demand zones, so we generated the boundary as convex hulls around the demand zones.
For Chicago, we only considered zones with at least \SI{30}{requests} to reduce the area and increase the density of the demand.
The area boundaries and zones for all areas are presented in Figure~\ref{fig:zones}.
Note that the DC zones have irregular shapes.
This is because the demand zones in DC are specified as Master Address Repository zones\footnote{Method of address standardization used in DC, \url{https://octo.dc.gov/node/715602}}. As we do not have access to boundaries for these zones, only their centroids, we generated the zone boundaries as Voronoi cells based on these centroids.

\begin{figure}
    \centering
    \begin{subfigure}[b]{0.49\linewidth}
        \includegraphics[width=\linewidth]{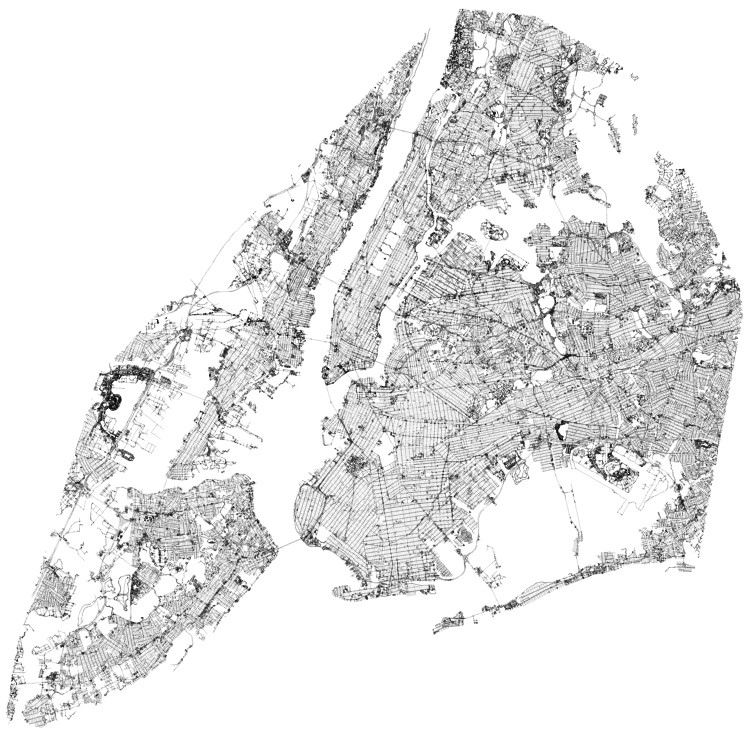}
        \caption{New York City} 
    \end{subfigure}
    \begin{subfigure}[b]{0.49\linewidth}
        \includegraphics[width=\linewidth]{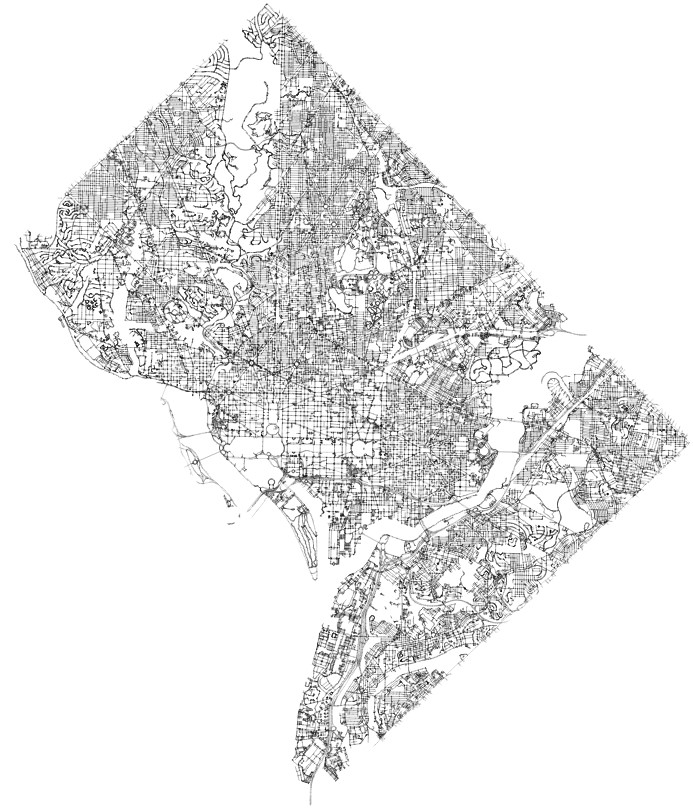}
        \caption{DC} 
    \end{subfigure}
    \caption{Example road networks}
    \label{fig:roads}
\end{figure}

The road networks were processed according to the steps described in the methodology section.
Only the roads within the area boundary (see Figure~\ref{fig:zones}) were used to generate the road network for each area. 
Two example road networks are visualized in Figure~\ref{fig:roads}.
The statistics for these road networks are summarized in Table~\ref{tab:roads}.
\begin{table}
    \centering
    \caption{Road network statistics.}
    \label{tab:roads}
    \footnotesize
    \settowidth\tymin{Manhattan}
    \setlength\extrarowheight{2pt}
    \begin{tabulary}{\linewidth}{LRRRR}
    Area & Node Count & Edge Count & Road length [km] & Area [km\textsuperscript{2}] \\
    \hline
    NYC & 113411 & 281278 & 27721 & 1508 \\
    Manhattan & 6382 & 13455 & 1329 & 87 \\
    Chicago & 152653 & 413830 & 31982 & 1004 \\
    DC & 33230 & 84788 & 5877 & 181 \\
    \end{tabulary}
\end{table}

\begin{figure}
    \centering
    \includegraphics[width=\linewidth]{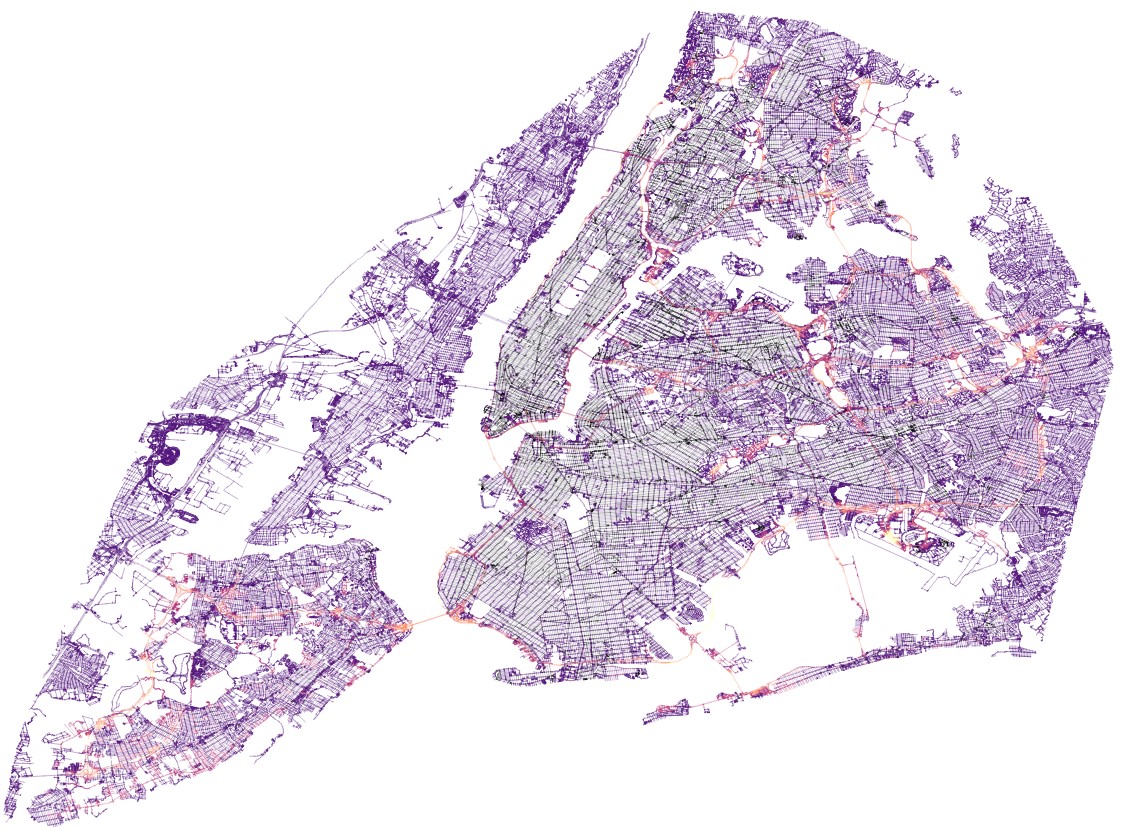}
    \caption{Speed map of New York City. The lighter color translates into higher speeds.}
    \label{fig:speeds}
\end{figure}

Each edge in the road network of all instances has speed associated with it.
The speed data were sourced from the Uber Movement dataset for the New York City and Manhattan areas. This data is visualized in Figure~\ref{fig:speeds}.
We lacked a data source with a similar level of detail for Chicago and Washington, DC, so we associated an average speed from the Uber Movement dataset with all edges.

Finally, by combining the road graph with the speed data and shortest path planner, we generated a distance matrix for each area that determines the travel time between any two locations (intersections in the road graph) and comprises the travel time model \(f_t\).
This matrix is enough to very quickly provide any necessary travel time information for DARP solvers and similar algorithms; the map itself is an intermediate product provided as supplemental material for visualization purposes and the user's convenience.

The dataset we present currently contains 96 instances, 24 for each area.
The parameters according to which these instances were generated are listed in Table~\ref{tab:instance_params}.
\begin{table}
    \centering
    \caption{Parameters for instance generation}
    \label{tab:instance_params}
    \begin{tabulary}{\linewidth}{LL}
    Start time & 2022-04-05 18:00:00 \\
    Duration [min] & 0.5, 1, 2, 5, 15, 30, 120, 960 \\
    Maximum delay ($ \Delta $) [min] & 3, 5, 10 \\
    Vehicle capacity ($ c $) [persons] & 4 \\
    Location & NYC, Chicago, Manhattan, DC
    \end{tabulary}
\end{table}
Note that instances with different parameters can be generated by following our methodology; this is just an example set with parameters set to values relevant (to our best knowledge) to MoD. 

\begin{figure}
    \centering
    \includegraphics[width=0.6\linewidth]{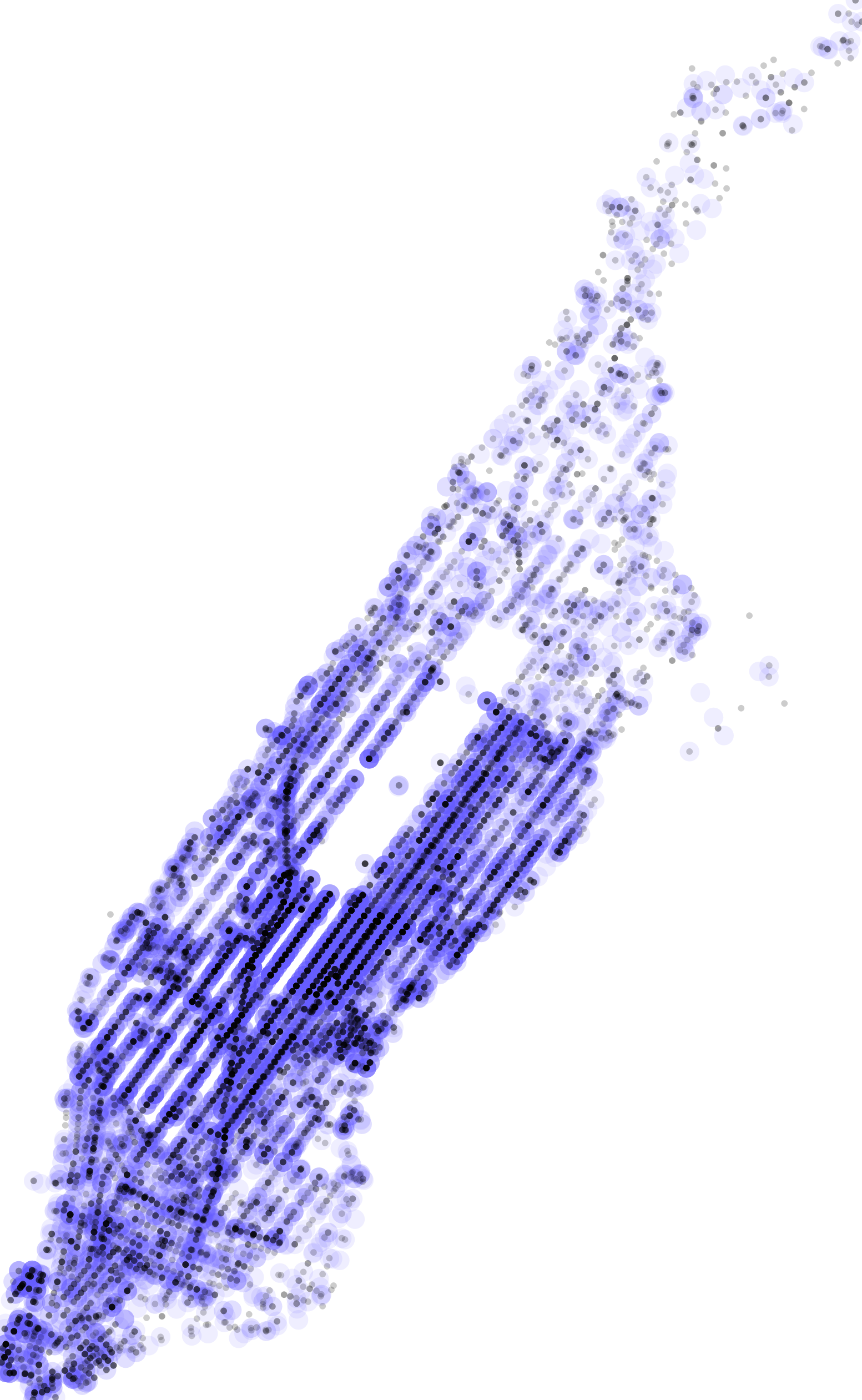}
    \caption{Example of a Manhattan instance. The purple areas mark demand: darker color translates to higher travel demand from the area. The Black circles mark vehicles at their initial positions. This particular instance contains \num{10362} requests spanning 30 minutes}
    \label{fig:instance}
\end{figure}

In addition to the travel time model in form of distance matrix \(f_t\), each instance contains a list of travel requests \(R\), a list of vehicles \(V\), and its configuration.
An example of the proposed instance is in Figure~\ref{fig:instance}.
The instances are complemented by additional meta-data, such as the timestamp of the start time of the demand, location name, road graph, travel speed data, and other supporting information useful for visualization and analysis.

\section{EXPERIMENTS}
We run a series of experiments to showcase the potential usage of our instances.
These experiments use two DARP solution methods to solve DARP problems defined by our instances.

\subsection{Solution Methods}
We evaluated two existing methods for solving the DARP problem:
\begin{itemize}
    \item the well know Insertion Heuristic (IH),
    \item and an optimal solution method, the Vehicle-group Assignment method (VGA)~\cite{alonso-moraOndemandHighcapacityRidesharing2017}.
\end{itemize}
IH is a standard heuristic for vehicle routing problems, including DARP.
Because of its good tradeoff between solution quality and computational requirements, it is frequently used as a default or baseline method~\cite{campbellEfficientInsertionHeuristics2004,kalinaAgentsVehicleRouting2015,bischoffCitywideSharedTaxis2017,fiedlerImpactRidesharingMobilityonDemand2018}.
Moreover, many metaheuristic methods use it to compute the initial solution~\cite{muelasVariableNeighborhoodSearch2013}.
The VGA method is an optimal solution method. 
Based on the existing research, one cannot expect it to solve all instances, especially those with large maximum time delays~\cite{fiedlerLargescaleOnlineRidesharing2022}.

\subsection{Implementation and Configuration}
We implemented both solution methods in C++ using  Gurobi solver\footnote{State-of-the-art commercial MILP solver, \url{https://www.gurobi.com/}} to compute optimal vehicle-group assignments in the VGA method.
We set the time limit for each experiment to \SI{24}{h} and ran each on AMD EPYC 7543 CPU with between \SI{1}{GB} and \SI{300}{GB} of memory and between 1 and 32 threads, depending on the location and solution method (VGA multi-threaded, IH single thread). 

\subsection{Results}
\begin{figure*}
    \centering
    \includegraphics[width=1\linewidth]{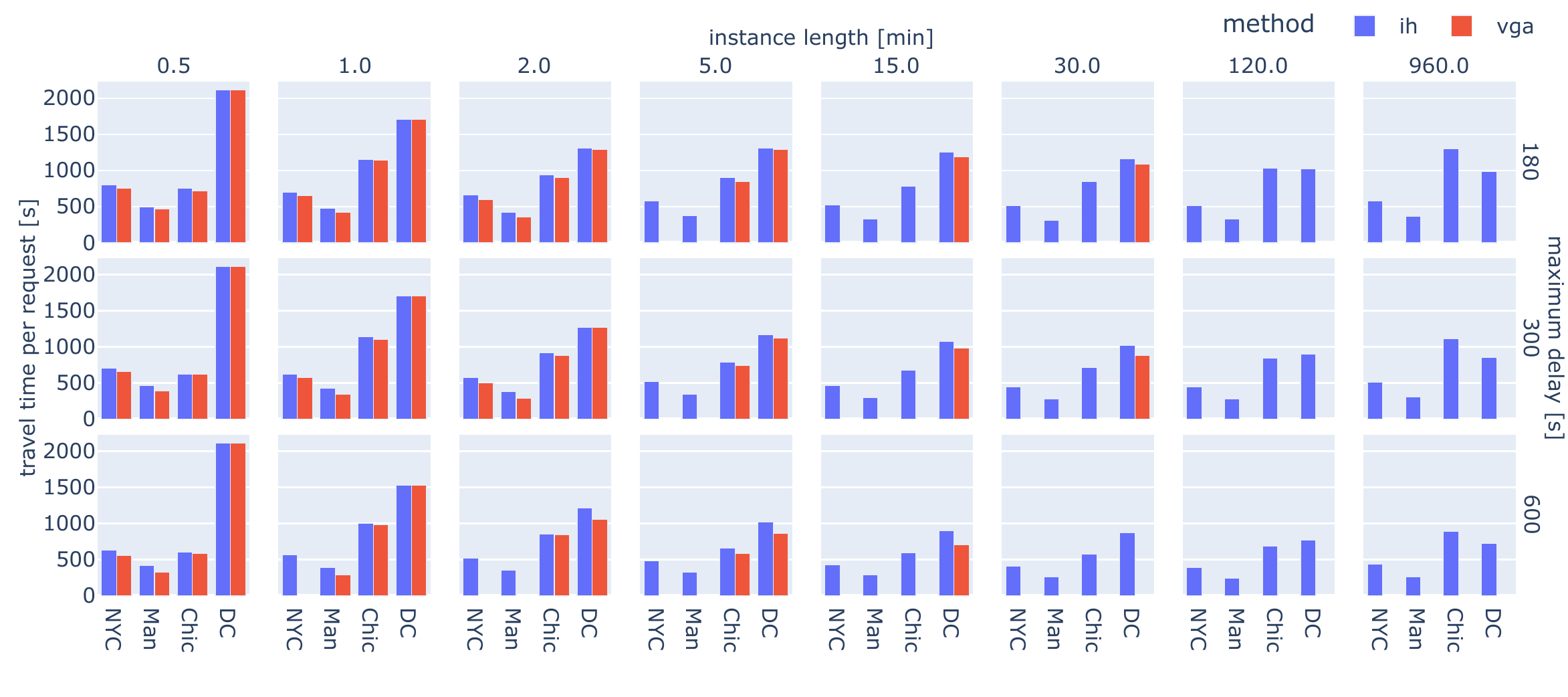}
    \caption{Average vehicle travel time per request. 
    }
    \label{fig:main_results}
\end{figure*}

We ran both methods on all instances.
You can inspect the full results as a part of the instance repository~(footnote 3 at page 1).
The main results showing the average cost (vehicle travel time) per travel request are in Figure~\ref{fig:main_results}.
As expected, the VGA method could not solve all instances within the time limit.
Another trend in line with previous work is that the cost decreases with the increasing max allowed delay.
The same is true for the instance duration.
All these trends correspond to findings from previous works~\cite{alonso-moraOndemandHighcapacityRidesharing2017,fiedlerLargescaleOnlineRidesharing2022}.

A more interesting comparison is that between different areas.
We can see that area is a more significant predictor of cost than instance duration or max delay.
One can argue that this is just an effect of different demand densities but a quick look at Table~\ref{tab:sources} does not support this simple conclusion. 
The Washington DC instances have greater average costs than the Chicago instances, while the demand density is four times greater.
When we compare the costs from New York and Manhattan instances, the less dense New York area indeed shows higher costs.
However, the cost difference is not that high if we consider that the demand density is more than ten times lower in NYC.
Both these comparisons suggest that there is no simple relation between the demand density and solution cost. 
Most likely, other more complex area properties, like those examined in Figure~\ref{fig:area_stats}, affect the solution quality as well.

\begin{figure*}
    \centering
    \includegraphics[width=1\linewidth]{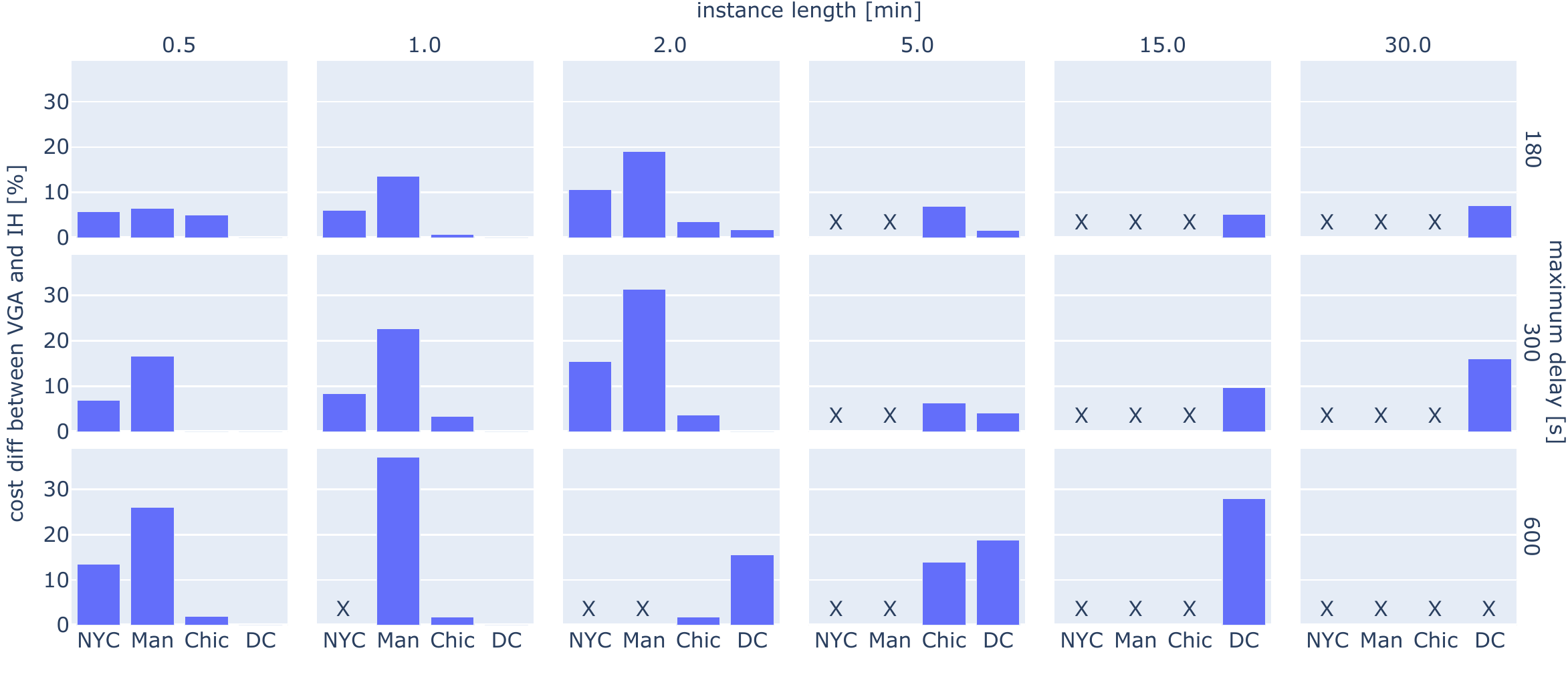}
    \caption{The increase of cost when solving by a suboptimal method (IH) relative to the optimal solution (VGA). 
    Instances with duration \SI{2}{h} and \SI{16}{h} are omitted.
    The crosses signalize missing values (missing results of the VGA method)
    }
    \label{fig:cost_ratio}
\end{figure*}

Another important question is how the area affects the cost of the heuristic solution compared to the optimal one. 
We can see that the difference is not consistent between areas.
Figure~\ref{fig:cost_ratio} brings more insight into this problem showing the relative difference between the optimal and the heuristic solution for each instance.
Again, we can see that the difference between optimal and heuristic solutions increases with instance flexibility (greater duration and maximum delay).
Nevertheless, the differences among areas are even greater, suggesting that area-specific properties are an important factor in determining the performance of heuristic methods.

\begin{figure}
    \centering
    \includegraphics[width=1\linewidth]{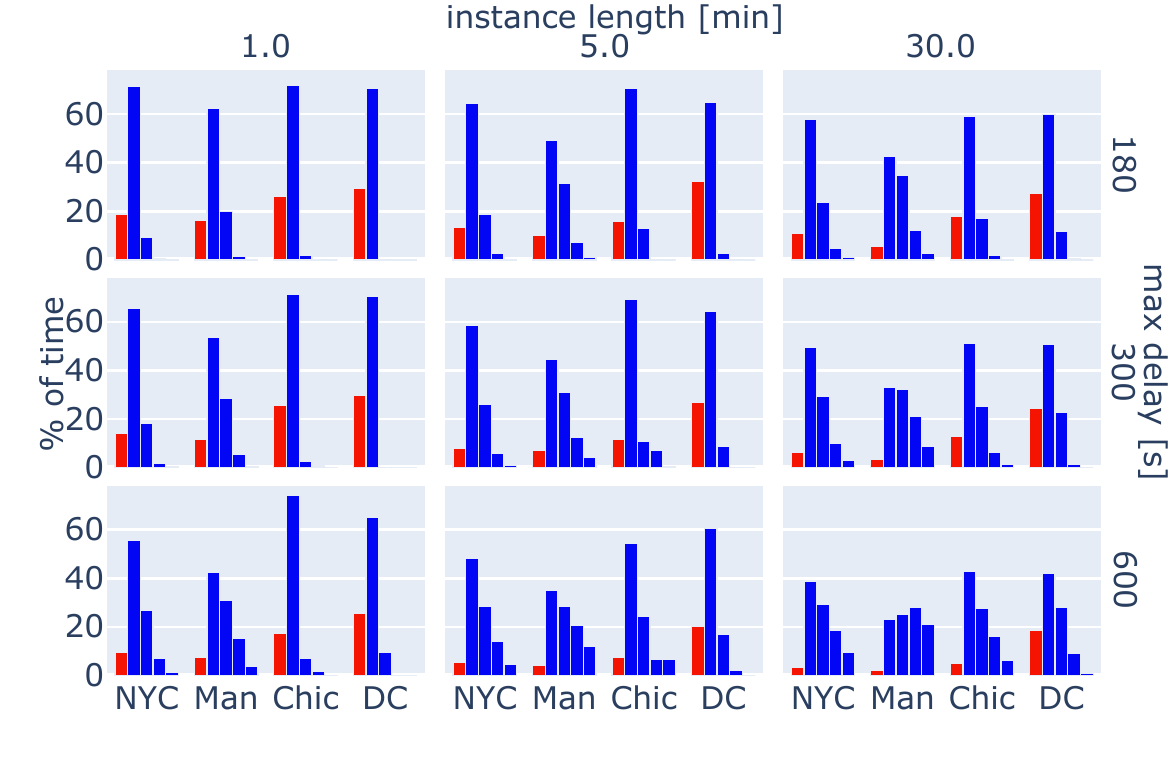}
    \caption{Occupancy histogram on selected instances when the problem is solved by the suboptimal method (IH). The red bars mark zero occupancy (empty vehicles), and the blue then starts from 1 on the left to 4 (vehicle capacity) on the right.}
    \label{fig:occupancy}
\end{figure}

For the final comparison, we show the histogram of occupancies on selected scenarios in Figure~\ref{fig:occupancy}.
Again, the occupancy is more impacted by the area than by the other parameters.
Manhattan has the highest average occupancy, and the lowest is in Washington DC instances.
But the instance length also has a considerable impact on occupancy. 
We can observe that the occupancy is low in the short instances, most likely due to the inability to match multiple requests in such a short time. 

\section{CONCLUSIONS}
In vehicle routing problems with time windows, and specifically dial-a-ride problems (DARP), significant efforts are dedicated to developing new methods that strike a better trade-off between solution cost and computational requirements.
With the emergence of new technologies and the digitalization of transportation, these problems have gained relevance in the context of large-scale mobility-on-demand (MoD) systems.
These systems are operated by transportation network companies, some of which already offer ride-sharing options. 
This is reflected in research where many works analyze the sharability and fleet-sizing and provide DARP solution methods for these systems.

To evaluate these solution methods, we need problem instances containing the travel demand, expected travel times, and vehicles.
However, the traditional DARP instances fall short when evaluating methods intended for large-scale MoD systems, even though the problem formulation remains largely identical.
Substantial differences in instance characteristics, such as demand density or time window sizes, are the reason.
Because of that, works that present solution methods for large-scale DARP usually use different problem instances. Those instances are, however, neither standardized nor publically available. Additionally, most of them are located in the Manhattan area, which can hardly be seen as a representative area due to its abnormally high demand density.
To make the matter even worse, the obfuscation of the publicly available datasets (including NYC) leads researchers to reuse the datasets released prior to the establishment of the privacy protection measures, which are now at least ten years old.

This research introduces a comprehensive collection of large-scale instances designed explicitly for evaluating solution methods tailored to large-scale MoD systems with ridesharing capabilities. 
We have carefully crafted instances with varying characteristics in four areas: New York City, Manhattan, Chicago, and Washington, DC.
All instances use actual demand data from these regions, including real travel times for New York City and Manhattan.
In total, the dataset contains 96 distinct problem instances.
With these instances, new DARP solution methods can be compared with the existing ones without reimplementing them.
Moreover, the methodology and implementation described in  this work can be used to generate additional instances tailored for representing current or future problems.

To showcase the intended usage, we evaluated all instances using two methods, the Insertion Heuristic (IH) and the optimal Vehicle-Group Assignment method.
Overall, the results align with findings from previous works: the solution cost per request is decreasing with both the instance duration and the maximum delay for request, and the IH suboptimality is mostly below \SI{20}{\percent}.
However, an interesting finding is the heavy dependence of all results on the characteristics of the specific areas. Thus, we conclude that evaluating new methods in multiple cities is crucial for methods intended for use within large-scale MoD systems. This contrasts with the current research practice.


Apart from its direct applicability to static DARP, our instances hold great potential for evaluating solution methods addressing online DARP and other MoD-related problems, such as fleet sizing, vehicle depot positioning, pricing, and future demand estimation.
Unlike classical DARP instances, which often rely on fictional data, the proposed instances are derived from real-life demand data and road networks and could be used to estimate the real-life performance of DARP solvers. 

In the future, we would like to generate more scenarios based on the feedback from the research community and evaluate multiple solution methods using our instances.





\bibliography{references-bibtex}

\end{document}